\documentclass{article}

\usepackage{PRIMEarxiv}

\usepackage[utf8]{inputenc} 
\usepackage[T1]{fontenc}    
\usepackage{hyperref}       
\usepackage{url}            
\usepackage{booktabs}       
\usepackage{amsfonts}       
\usepackage{nicefrac}       
\usepackage{microtype}      
\usepackage{lipsum}
\usepackage{fancyhdr}       
\usepackage{graphicx}       
\graphicspath{{media/}}     
\usepackage{tabularx}
\usepackage{booktabs}
\usepackage{makecell}
\usepackage{siunitx}
\usepackage{amsmath}
\usepackage{multirow} 
\usepackage{float}
\usepackage{authblk}
\title{

Integrating Symbolic Neural Networks with Building Physics: A Study and Proposal
 
}

\author[1, 3]{\textbf{Xia Chen} *}
\author[2, 3]{\textbf{Guoquan Lv} *}
\author[3]{\textbf{Xinwei Zhuang} }
\author[3]{\textbf{Carlos Duarte} }
\author[3]{\textbf{Stefano Schiavon} }
\author[1]{\textbf{Philipp Geyer} }

\affil[1]{Sustainable Building Systems, Leibniz University Hannover, Hannover, Germany\\
\texttt{xia.chen/philipp.geyer@iek.uni-hannover.de\\}}

\affil[2]{Department of Architecture, Zhejiang University, Hangzhou, China\\
\texttt{guoquanlv@intl.zju.edu.cn\\}}
  
\affil[3]{Center for the Built Environment, University of California, Berkeley, Berkeley, USA\\
\texttt{xinwei\_zhuang/cduarte/schiavon@berkeley.edu}  }

\begin{document}
\maketitle

\begin{abstract}
Symbolic neural networks, such as Kolmogorov–Arnold Networks (KAN), offer a promising approach for integrating prior knowledge with data-driven methods, making them valuable for addressing inverse problems in scientific and engineering domains. This study explores the application of KAN in building physics, focusing on predictive modeling, knowledge discovery, and continuous learning. Through four case studies, we demonstrate KAN's ability to rediscover fundamental equations, approximate complex formulas, and capture time-dependent dynamics in heat transfer. While there are challenges in extrapolation and interpretability, we highlight KAN's potential to combine advanced modeling methods for knowledge augmentation, which benefits energy efficiency, system optimization, and sustainability assessments beyond the personal knowledge constraints of the modelers. Additionally, we propose a model selection decision tree to guide practitioners in appropriate applications for building physics.
\end{abstract}

\begin{figure}[h]
    \centering
    \includegraphics[width=0.9\linewidth]{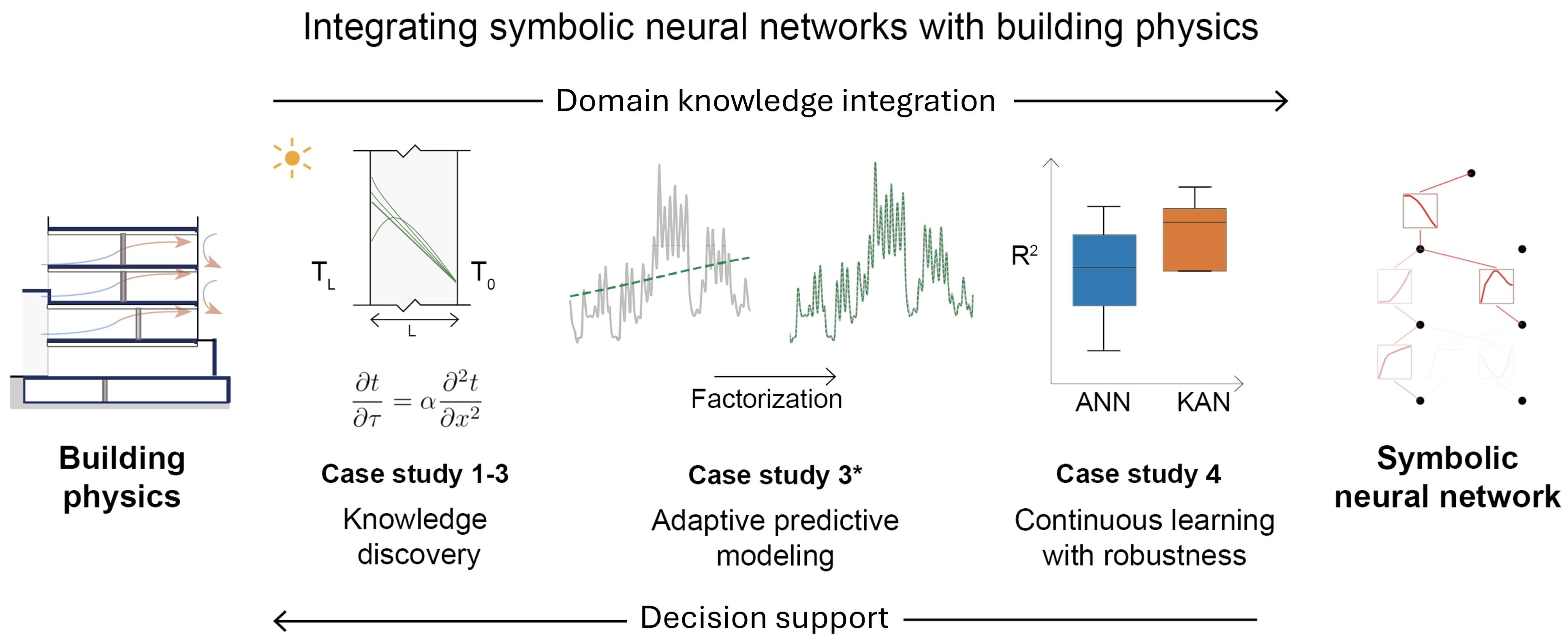}
    \label{fig:Gr_Ab}
\end{figure}

\section{Introduction}
\label{sec:Introduction}


The emergence of data-driven methods and their end-to-end modeling behavior without the comprehensive, explicit system understanding required has transformed problem-solving approaches across various engineering fields. Engineers use these techniques for pattern-matching capabilities and efficient solution finding. However, in the field of building physics, achieving accurate results is only part of what is relevant for designers and researchers. The need for explainability and reasoning is crucial for optimization and control. In these contexts, understanding the underlying factors holds equal importance to the solutions themselves\cite{marcondes2024back}. This dual requirement for accuracy and interpretability presents a unique challenge in applying data-driven methods to building physics.

While statistical models provide foundational approaches for numerical modeling and forecasting, machine learning (ML) methods further boost application effectiveness. Starting from logistic regression or ridge regression, ARIMA \cite{yildiz2017review}, to more advanced models like kernel machines, random forests \cite{sun2021machine}, and gradient boosting machines \cite{touzani2018gradient}, these models offer robust performance in non-linear regression tasks and have become integral in engineering application, such as personal comfort model development \cite{kim2018personal},  structural engineering \cite{salehi2018emerging}, and urban building energy modeling \cite{MARL}. Beyond traditional ML models, Artificial Neural Networks (ANN) or Multi-Layer Perceptrons (MLPs) have been widely adopted due to their ability to approximate complex functions through multiple layers of non-linear activations and linear transformations. Variants like Recurrent Neural Networks (RNNs) and Long Short-Term Memory (LSTM) networks are particularly effective for sequential data capture temporal dependencies. Their application and refined models have shown success in various predictive tasks such as building load forecasting, energy use prediction \cite{wang2020building}, fault detection in energy systems \cite{ZHANG2023100235}, while Physics-Informed Neural Network (PINN) \cite{raissi2019physics, cuomo2022scientific} extends from MLPs with prior domain knowledge as constraints to address inverse physics problems, showcasing the versatility of neural network in approaching solutions efficiently.

Two efforts are being pursued in the data-driven methods applied to building physics. The first focuses on efficiently capturing the patterns behind the data by incorporating domain knowledge \cite{von2021informed} to improve predictive accuracy \cite{chen2022physical, cai2021physics, gokhale2022physics}. Techniques in this line aim to refine model architectures \cite{chen2024utilizing} or algorithms \cite{chen2022hybrid} to better reflect the physical realities of the modeled systems. The second effort focuses on the interpretability of the models and the ability to make decisions. This involves developing models and methods that predict outcomes and provide insights into the underlying mechanisms and relationships within the data \cite{cuomo2022scientific}. While significant progress has been made in the first area, the second remains under development, highlighting the need for models that can offer both accuracy and interpretability. Data-driven models in building physics often face challenges related to interpretability and generalization \cite{chen2023interpretable}. For example, in design, retrofit, or control optimization scenarios, when transferring trained models from one building to another, it can lead to significant drops in model performance due to wide variations in structure organization and material combinations \cite{pinto2022transfer}.

We observed a need for predictive modeling, moving from prior knowledge integration to building models directly from the data for knowledge discovery \cite{cios2012data, chen2023pathway}. One of the promising paths is using symbolic regression (SR) and symbolic optimization (SO) \cite{sun2019data, tenachi2023deep, angelis2023artificial}, differently from traditional regressions, these approaches do not assume a specific form, like linear or quadratic, before searching a solution. SR methods promise to advance predictive modeling and provide a framework for knowledge discovery \cite{alshahrani2017neuro}, this dual capability makes them a valuable tool for addressing the multifaceted challenges in building engineering and energy-efficient design, where both performance optimization and a deep understanding of underlying mechanisms from data and phenomenological analysis are essential.

In this paper, we examine one of the most recent methods in the SR family, Kolmogorov-Arnold Network's (KANs) \cite{liu2024kan} ability to bridge the gap between data-driven models and domain-specific insights, investigating its capabilities and constraints. By leveraging the strengths of both human expertise and advanced machine learning systems, we aim to achieve solutions that are not only more reliable and accurate but also interpretable and, therefore, actionable. Furthermore, we propose a decision tree for selecting the right tools given the problem at hand. 


\subsection{Introduction to Kolmogorov-Arnold Networks}
\label{subsec:KAN_intro}

Kolmogorov-Arnold Networks is a class of neural networks inspired by the Kolmogorov-Arnold representation theorem \cite{kolmogorov1961representation}, which posits that any multivariate continuous function can be represented as a superposition of continuous functions of a single variable and addition. KANs provide a powerful framework for decomposing complex multivariate functions into simpler, more manageable and interpretable components. Unlike traditional ANNs or MLPs that have fixed activation functions on nodes, KANs achieve higher accuracy and interpretability with fewer parameters through spline-based activation functions \cite{liu2024kan}. KANs tend to outperform MLPs in data fitting and solving partial differential equations \cite{liu2024kan}. These features are crucial in fields like building physics, especially when using functions with polynomial or exponential terms. For example, heat transfer equations use exponentials. 

Alternative to traditional ANNs, the adaptable architecture of KANs allows for the effective handling of diverse data types and tasks, ranging from graph processing to time series forecasting \cite{kiamari2024gkan,vaca2024kolmogorov}. Furthermore, their potential in symbolic regression enables the discovery of interpretable formulas \cite{schmidt2009distilling}, which is particularly valuable in building physics. Yet, as the complexity of the modeled system increases, the resulting symbolic expressions can become challenging to interpret, necessitating further research into simplifying these outputs while retaining accuracy \cite{hou2024survey}.

\subsection{Application of KANs in Building Physics}
\label{subsec:KAN_building_physics}

Building upon the above-mentioned features, the integration of KANs into building physics provides a potential approach to address domain-specific challenges, such as thermal performance analysis and energy efficiency optimization, which face unique difficulties that distinguish it from other fields:

\begin{itemize}
    \item Building environments change frequently, encountering new scenarios and operational shifts. For instance, the recent surge in teleworking inherent from the pandemic impact has dramatically altered building occupancy patterns \cite{ouf2019quantifying}. Concurrently, extreme weather events continue to reshape environmental conditions \cite{prabowo2023continually}. These rapid changes demand models with robust adaptability and extrapolation capabilities, models that can adjust to scenarios that deviate significantly from historical data. 
    
    \item Data collection in building physics is often expensive and time-consuming \cite{wang2022data}, necessitating efficient utilization of the limited datasets. This demands models capable of small-sample learning and effective generalization from sparse data.
    
    \item While accuracy in the prediction is important, the ability to generalize across diverse building types, programs and climates is often more important than achieving high accuracy on a specific dataset \cite{pinto2022transfer, liu2021transfer} for predictive modeling. 

    \item In design or retrofitting stages, it is essential to properly dimension the building energy system without underestimating the peak performance and demand from historical behavior \cite{alduailij2021forecasting}. 
    
    \item Finally, the black-box nature of many modeling approaches is insufficient in the domain decision-making process. Understanding the underlying physical principles is essential for engineers and policymakers to make informed decisions about building design and operation \cite{ali2020data, manfren2022data}. 
\end{itemize}

At its core, addressing these challenges normally requires the model not only to capture the pattern behind the data implicitly but also to express and model the system in an explicit, interpretable way. On the one hand, such explicit modeling ability is essential to allow the model to perform or even extrapolate in various conditions because it allows for intervention or cross-validation via our domain knowledge; on the other hand, the complexity of KANs can hinder timely application, and while they offer high accuracy, their symbolic outputs often become difficult to interpret as system complexity increases, which raise the question of the model selection problem between KANs and MLPs. We aim to explore the application of KANs in building physics, focusing on: (a) Investigate how KANs can be employed to handle complex, non-linear relationships for forecasting and optimization across diverse building scenarios (for example, what happens if there is an operational shift, like starting the building earlier in the day); (b)Employ KANs to uncover the physics at the base on limited datasets. (c) Evaluate KANs' capability to generalize from small datasets to a wide range of building types, programs and climates by incorporating continuous data inputs. 

Additionally, we summarize our discoveries into a decision tree that guides the user in the selection of KANs vs MLP or other ANNs vs traditional methods. 

The rest of this paper is organized as follows: Section \ref{sec:case_studies} describes the basic landscape of setup for numerical experiments, selecting three common thermal calculation models to testify KAN's knowledge extraction capacities; interestingly, we discovered a knowledge-integrated approach for KAN's performance improvement. Followed with an additional case to examine the model robustness against data sparseness, continuous learning ability, and capacity in handling extreme value capture performance between MLP and KAN. Section \ref{sec:results_and_discussion} presents the results and discusses the interpretability of KANs in this domain. Finally, Section \ref{sec:Framework} proposes a decision tree for algorithm selection.

\section{Description of Case Studies}
\label{sec:case_studies}
We present a series of case studies designed to evaluate the performance of KANs in solving inverse problems pertinent to building thermal physics. First, we examine three case studies that utilize typical building performance models that progressively increase in complexity, allowing for a comprehensive assessment of KAN's capabilities across a range of scenarios critical to understanding and optimizing building thermal performance. Subsequently, we introduce an additional case study that compares KAN's capacity with that of MLPs, employing a large-scale building dataset.


The three case studies share a common framework rooted in the fundamental principle physics of heat transfer through building envelopes. We consider one-dimensional heat flow through a homogeneous wall, and assume constant thermal properties of the wall material, such as thermal conductivity and thermal diffusivity. The boundary conditions are specified as temperatures at the wall surfaces, simulating the interface between the building interior, the wall, and the outdoor environment. Each case has analytical solutions, and these numerical solutions serve as ground truth, enabling the evaluation of KAN's performance. The major objective across all three cases is to evaluate KAN's ability to discover the underlying physical relationships from data without prior knowledge of the "real" equations governing the heat transfer process.


Table \ref{tab:case_parameters} summarizes the key parameters for each case:

\begin{table}[h]
\centering
\caption{Summary of case study parameters with increasing complexity.}
\label{tab:case_parameters}
\begin{tabularx}{\textwidth}{@{}l>{\centering\arraybackslash}m{4cm}>{\centering\arraybackslash}m{4cm}>{\centering\arraybackslash}m{4cm}@{}}
\toprule
 & Case 1: Steady-State & Case 2: Transient & Case 3: Dynamic \\
\midrule
Time Dependence & Steady-state & Time-dependent & Time-dependent with periodicity \\
\midrule
Description & Simplified wall in a controlled environment, typical for laboratory or idealized simulations. & Simulates the wall's response to sudden temperature changes. & Represents real-world walls exposed to diurnal temperature cycles. \\
\midrule
Analytical Solution & Linear Equation, eq. (1) & Error Function Solution,  eq. (4) & Harmonic Method, eq. (5) 
Response Factors Method, eq. (6)\\
\bottomrule
\end{tabularx}
\end{table}

\subsection{Case Study 1: Steady-State Heat Conduction}
\label{subsubsec:case_study_1}

The first case study examines steady-state heat conduction through a homogeneous wall \cite{som2008introduction}, representing the simplest scenario of heat transfer in building envelopes. We consider a wall with a thickness of 0.24 m, exposed to constant temperatures of 20°C on the interior surface and 40°C on the exterior surface. This scenario mimics a simplified building wall under steady-state conditions in a controlled environment, such as in laboratory settings or assessment of the thermal resistance. The steady-state condition allows us to focus on KAN's ability to capture spatial variations in temperature without the added complexity of time dependence.

The analytical solution for this case is a simple linear equation. The simplicity of this case also serves as a foundation for understanding KAN's behavior as we progress to more complex scenarios.
\begin{equation}
\label{wall steady conduction}
T(x) = T_0 + \frac{T_L - T_0}{L} x
\end{equation}

where, $T_0$ and $T_L$ represent the temperatures at the wall at positions $x=0$ and $x=L$, respectively.

\subsection{Case Study 2: Transient Heat Conduction}
\label{subsubsec:case_study_2}

Building upon the first case, the second study introduces the temporal dimension to examine transient heat conduction through soil \cite{incropera2007fundamentals}. This case challenges KAN to capture both spatial and temporal dynamics of heat flux and resulting temperature, representing a significant increase in complexity. The semi-infinite medium model is commonly used for calculating heat transfer through soil and building ground. The thermal diffusivity of the soil medium is set at \(2.5 \times 10^{-6} \, \text{m}^2/\text{s}\) according to the median value of published data \cite{arias2015determining, zhao2022assessing}. Initially, the medium is at a uniform temperature of 20°C. We then apply a step change in temperature to 40°C at the external surface. This scenario simulates the thermal response of the soil to a sudden change in external temperature, as might occur at sunrise or with the onset of a heatwave.

The analytical solution for this case involves the complementary error function (\(\text{erfc}\)), capturing the nonlinear behavior of heat diffusion through the wall over time. This nonlinearity presents a more significant challenge for KAN, testing its ability to discover and represent complex physical relationships.

The complementary error function (\(\text{erfc}\)) is a mathematical function related to the error function (\(\text{erf}\)), which is widely used in probability, statistics, and partial differential equations to describe diffusion processes. The \(\text{erfc}\) function is defined as:

\begin{equation}
\text{erfc}(x) = 1 - \text{erf}(x)
\end{equation}

where

\begin{equation}
\text{erf}(x) = \frac{2}{\sqrt{\pi}} \int_0^x e^{-t^2} \, dt
\end{equation}

In the context of heat conduction, the \(\text{erfc}\) function effectively models the temperature distribution within a material over time, providing a description of how heat diffuses through the material after a sudden change in surface temperature.

In our study, the complementary error function is not part of the original set of operators available within KAN. To address this, we manually encoded the \(\text{erfc}\) function into KAN to ensure it could accurately fit the observed data. This extension allows KAN to leverage the mathematical properties of the \(\text{erfc}\) function, enabling it to capture the transient heat conduction behavior more precisely. By incorporating this additional operator, we enhance KAN's ability to model complex physical phenomena. The analytical solution formula is as follows. 
\begin{equation}
\label{transient}
T(x,\tau) = T_0 + (T_1 - T_0) \operatorname{erfc}\left(\frac{x}{2\sqrt{\alpha \tau}}\right)
\end{equation}

where, $T_0$ represents the initial temperature of the entire object, $T_1$ represents the sudden temperature change at the boundary of the object, $\alpha$ refers to the thermal diffusivity coefficient.

\subsection{Case Study 3: Dynamic Heat Transfer with Periodic Boundary Conditions}
\label{subsubsec:case_study_3}

The third case study represents the most complex and realistic scenario, simulating transient heat transfer through a concrete wall with a time-varying external temperature. This case more closely mimics real-world conditions where building walls are exposed to diurnal temperature variations, accounting for solar heat gain patterns and daily temperature cycles. This periodic boundary condition introduces a new level of complexity, challenging the method to accurately capture periodic thermal behaviors crucial for practical building energy analysis.

We consider a concrete wall with a thickness of 0.24 m. For this one-dimensional dynamic heat transfer, it can be solved by harmonic method \cite{ricciu2019thermal, varela2012harmonic}. The harmonic method decomposes the approximate periodic meteorological data through a series of sine (or cosine) functions. By using the wall's transfer function, derived from performing a Laplace transform on the wall's heat conduction differential equation, the attenuation and delay of each term in the series can be calculated \cite{building_thermal_processes_1986}. And then the heat flow (HF) on the interior side of the wall can be calculated by eq.(\ref{HF_harmonicmethod})
This yields an analytical solution that is represented by a combination of a sine series term and a linear formula, providing a high-fidelity representation of the complex heat transfer process (the detail can be found in Appendix \ref{app:case3_setup}). The external temperature variation is represented by the sol-air temperature \cite{OCallaghan1977}, calculated from real meteorological data. 

\begin{equation}
\label{HF_harmonicmethod}
HF(\tau) =  K\left[t_0 - t_{in} + \frac{h_{in}}{K} \ \sum_{n=1}^N \frac{A_{n}}{v_n} \sin(\omega_n \tau + \phi_n - \psi_n)\right]
\end{equation}

Where \( K \) is heat transfer coefficient of the wall, \( t_0 \) is average outdoor temperature, \( t_{\text{in}} \) is average indoor temperature, \( h_{\text{in}} \) is convective heat transfer coefficient on the indoor side, \( v_n \) is attenuation rate of the wall for the \( n \)-th harmonic component of the heat flow, \( A_n \) is amplitude of the \( n \)-th harmonic, representing the impact magnitude of this frequency component on temperature variations. \( \sin(\omega_n \tau + \phi_n - \psi_n) \) represents the sinusoidal waveform of the \( n \)-th harmonic, \( \omega_n \) is angular frequency of the \( n \)-th harmonic, \( \phi_n \) is phase of the \( n \)-th harmonic, \( \psi_n \) is phase delay caused by the thermal characteristics of the wall for the \( n \)-th harmonic.

The heat flow on the interior side of the wall can also be determined using the response factor method \cite{haghighat1992determination}. 
The principle involves performing a Laplace transform on the heat conduction equation of the wall, and then discretizing the outdoor temperature profiles for the analysis.  Then, by utilizing the residue theorem \cite{Stephenson1971}, we can solve for the response factors \( Y(j) \) to quantify the impact of each time node of outdoor temperature \( t_o(\tau-j) \) on heat flow, thereby determining the dynamic heat flow. The calculation can be represented by the following equation:

\begin{equation}
HF(\tau) = \sum_{j=0}^{\infty} Y(j) t_o(\tau-j) - K t_{in},
\end{equation}

This case tests KAN's ability to handle complex and changing boundary conditions and discover relationships in highly dynamic thermal systems. To allow for reproducing our results, we shared more details in Appendix \ref{app:reproducibility}.

\subsection{Case Study 4: Performance Comparison with MLPs}

In the previous cases, we explored KAN's performance in terms of knowledge extraction capabilities. In this case, our focus shifts to the performance comparison between KANs and MLPs. In this case, we are more concerned about the KANs' robustness against data sparseness, their capability of continuous learning, and the ability to capture extreme values.

The data for this case study are taken from a dataset originally designed to evaluate model robustness against sparse data and transfer learning capabilities \cite{chen2024utilizing}. We utilize data from a roof, in predicting heat flow time series based on various environmental and structural factors. The dataset includes parameters such as temperature, dew point, humidity, pressure, area, thermal transmittance (U-value), and heat capacity. The study aims to evaluate the effectiveness of these two approaches in capturing complex relationships within building thermal dynamics. The benchmark results from the original study indicate difficulty in accurately modeling roof temperature dynamics due to these slower response times. The dataset is available online \cite{Chen2024Knowledge-BasedSystems}. For a detailed dataset description and evaluation, we point to the original study \cite{chen2024utilizing}.

The case is characterized by extreme sparsity and so it can be used to test models' abilities to make accurate predictions with minimal input. This setup simulates real-world scenarios where data acquisition is challenging or incomplete, and models must generalize effectively from limited observations. The roof component's thermal response serves as a challenging test case, with its slower heat transfer rates requiring models to accurately capture the lagged response to external temperature fluctuations.

We compare KANs with MLPs based on: (a) their ability to learn from data. This aspect assesses each model's capability to adapt and improve over time as more data becomes available. We divided the dataset into three equal parts (Task 1, Task 2, and Task 3). We implemented a sequential learning scenario where models were trained on these tasks in succession, with varying data availability represented by different subsample rates (100\%, 50\%, 25\%, and 10\%). (b) Model robustness. This criterion evaluates how well each model can maintain performance levels when trained on highly sparse datasets. We trained both KAN and MLP models on progressively smaller subsets of the data, using subsample rates of 100\%, 50\%, 25\%, 10\%, and 5\% of the original dataset. The final data sample could be lower than the number of buildings included in the dataset, which would represent the scenario of data inconsistency and scarcity. (c) Ability to capture extreme values. This criterion measures each model's ability to predict rare or extreme events. To assess the models' ability to predict rare events (for example, a heatwave), we focused on their performance in capturing the top 25\%, 10\%, and 5\% of the extreme temperature variations and peak energy loads. 
For detailed information on the KAN implementation, including software libraries, architecture, and training process, please refer to Appendix \ref{app:kan_implementation}. For detailed experimental setup, methodology, and results, please refer to Appendix \ref{app:implementation}.

\section{Results and Discussion}
\label{sec:results_and_discussion}

This section presents the outcomes of our four case studies and provides a detailed analysis of KAN's performance in each scenario. We begin with a summary of quantitative results, followed by a discussion of each case study and a synthesis of our findings.

\subsection{Analysis of Results}
\label{subsec:analysis_of_results}

\begin{figure}[h]
    \centering
    \includegraphics[width=1\linewidth]{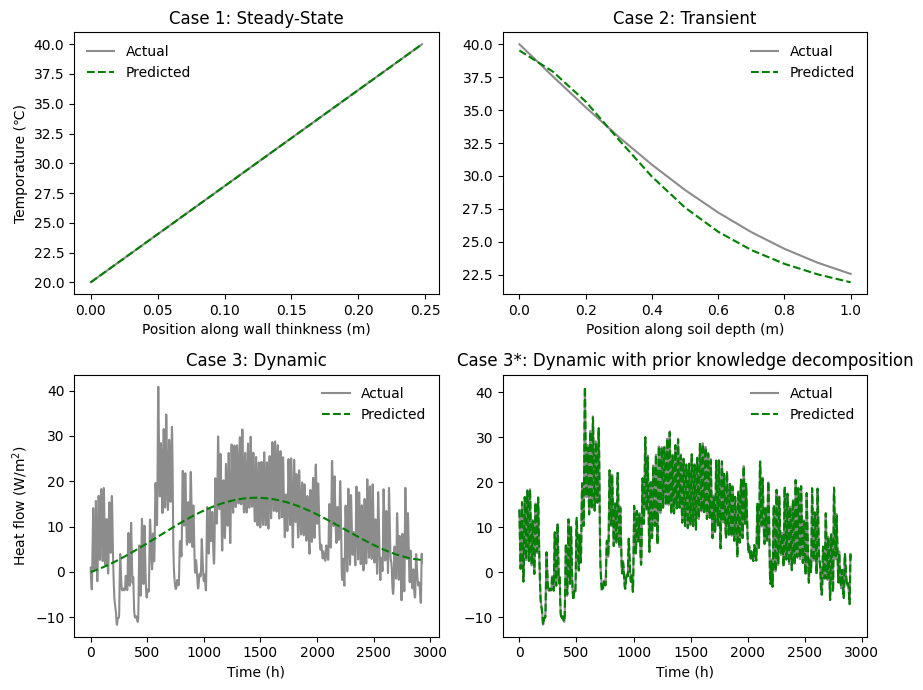}
    \caption{KAN prediction results across different case studies. Case 1: steady heat transfer in walls; Case 2: transient heat transfer in semi-infinite soil; Case 3: periodic dynamic heat transfer in walls; Case 3*: periodic dynamic heat transfer in walls with prior knowledge decomposition. }
    \label{fig:Re_Ca}
\end{figure}

Table \ref{tab:quantitative_results} summarizes the key performance metrics for all three case studies, as their predictive result comparison is illustratively presented in Figure \ref{fig:Re_Ca}.  In the context of this study, the complexity index that we estimated refers to the number of distinct terms or components present in an equation, including mathematical operations, functions, and constants. Lower complexity generally suggests a more straightforward and interpretable model, whereas higher complexity may indicate a better fit at the expense of model transparency and potential overfitting.

\begin{table}[h]
\centering
\caption{Comparison of KAN performance with analytical solutions across case studies with varying complexity levels}
\label{tab:quantitative_results}
\resizebox{\textwidth}{!}{%
\begin{tabular}{ccccc}
\toprule
Case & Complexity & Analytical Solution & KAN Generated Formula & R² \\
\midrule
1 & 2 & $T(x) = T_0 + \frac{T_L - T_0}{L} x$ & $T(x) = T_0 + \frac{T_L - T_0}{L} x$ & \textbf{1.00} \\
\midrule
2 & 5 & Error Function Solution, eq. (4) & Multiple $erfc$ terms, eq. (8)& \textbf{0.99} \\
\midrule
3 & 12 & Harmonic Method Equation, eq. (5) & Series with $sin$ terms, eq. (9)& \textbf{0.20}\\
\midrule
3* & 12 & Response Factors Method Equation, eq. (6)& Series with linear terms, eq. (10) & \textbf{0.99} \\
\bottomrule
\end{tabular}%
}
\vspace{10 pt} 
\raggedright 
\small{Note: 3* Results when we have a prior knowledge decomposition.}
\end{table}

The results highlight KAN's ability to adapt its complexity based on the requirements of the physical model it attempts to learn. Case 1, being a simple linear heat conduction model, shows both a low complexity and a high R-squared value, indicating a perfect match with the analytical solution. Cases 2 and 3, involving more dynamic and complex boundary conditions, demonstrate an increase in formula complexity, which corresponds with KAN's capability to handle more sophisticated scenarios. The low R-squared in Case 3, however, suggests potential issues with model fit or stability in highly dynamic environments.

\subsubsection{Case Study 1: Steady-State Heat Conduction}
\label{subsubsec:case1_analysis}

In the steady-state heat conduction case, KAN demonstrated exceptional performance, successfully rediscovering the linear relationship with remarkable accuracy (R² = 1.00). The learned formula perfectly matches the analytical solution:

\begin{equation}
T = 83.33\cdot x + 20
\end{equation}

This result is particularly noteworthy for two reasons. Firstly, the coefficient (83.33) almost perfectly matches the expected slope of $(T_L - T_0) / L = (40 - 20) / 0.24 \approx 83.33$. Secondly, the intercept (20) is virtually identical to the expected $T_0$ value of 20°C. This level of precision demonstrates KAN's ability to not only capture the linear nature of steady-state heat conduction but also to accurately determine the specific parameters of the system.

Figure \ref{fig:Re_Ca}, Case 1 visually confirms the performance of KAN in this case. The predicted temperature distribution (blue solid line) perfectly overlaps with the actual distribution (orange dashed line), showing a linear increase in temperature from 20°C at $x = 0 m$ to 40°C at $x = 0.24 m$. This performance in the simplest case study serves as a  validation of KAN's ability to discover basic physical laws from data.

\subsubsection{Case Study 2: Transient Heat Conduction}
\label{subsubsec:case2_analysis}

In the more complex scenario of transient heat conduction, KAN adeptly approximated the nonlinear heat transfer behavior, achieving high accuracy (R² = 0.99). The derived formula, more complex than the concise analytical solution represented by eq.(\ref{transient}), poses challenges in interpreting the physical significance of each term.

\begin{equation}
\begin{split}
T = & -10.79 \cdot \text{erfc}\left(-1.75 \cdot \text{erfc}\left(-0.65 \cdot \text{erfc}(12.34) - 1.77 \cdot \text{erfc}(1.32 - 13.44x_1) - 0.54\right)\right. \\
    & \left.- 1.79 \cdot \text{erfc}\left(0.01 \cdot \text{erfc}(24.36) - 0.98 + \frac{1.11}{\sqrt{x_1 + 0.18}}\right) + 4.44\right) + 40.61 \\
    & - \frac{0.03}{\sqrt{1 - \frac{0.01}{\sqrt{-0.01 \cdot \text{erfc}(24.36) + 1 - \frac{0.42}{\sqrt{x_1 + 0.18}}}} + \frac{0.01}{\sqrt{0.36 \cdot \text{erfc}(12.34) + \text{erfc}(1.32 - 13.44x_1) - 0.05}}}}
\end{split}
\end{equation}

Nevertheless, it captures the essence of the temperature distribution. The complementary error function (erfc) was manually encoded into KAN as an additional operator to ensure it could accurately fit the observed data. This demonstrates KAN's flexibility in incorporating domain-specific functions. The increased complexity of the formula (10 terms) compared to Case 1 reflects the more intricate nature of transient heat transfer. Despite this complexity, the high R² value indicates that KAN's predictions closely match the true temperature distribution across both space and time.

It's worth sharing that we ran the KAN algorithm multiple times, with R² values ranging from 0.92 to 0.99. The formula presented here represents one of the high-accuracy results. The generated formulas were not identical across runs, highlighting the stochastic nature of the formula discovery approaching process.

Figure \ref{fig:Re_Ca}, Case 2, visually confirms the high performance of KAN in this case. The predicted temperature distribution (blue solid line) closely follows the actual distribution (orange dashed line), showing a nonlinear decrease in temperature from left to right. The plot demonstrates KAN's ability to capture the transient behavior of heat conduction, with only slight deviations from the actual values, particularly near the boundaries. However, the structure of KAN's formula differs from the analytical solution, featuring nested erfc functions and additional terms involving square roots. While KAN's formula accurately predicts the temperature distribution, its structure may not directly reveal the underlying physical principles as clearly as the analytical solution. 

\subsubsection{Case Study 3: Dynamic Heat Transfer with Periodic Boundary Conditions}
\label{subsubsec:case3_analysis}

In the most complex case study, involving dynamic heat transfer with periodic boundary conditions, KAN demonstrated its ability to capture some of the spatial and temporal aspects of heat transfer. The discovered formula  (for the complete expression, see\ref{Case Study Results}) is:

\begin{equation}
HF(\tau) = \sum_{n=1}^{N} a_n \cdot \sin\left(b_n \cdot \tau + c_n\right)+D
\label{eq:HF_formula1}
\end{equation}

KAN successfully identified the periodic nature of the problem (it used the sine functions in the formula). This aligns with the sinusoidal external temperature variation and the expected wave-like behavior of heat transfer in this case. Nevertheless, the complexity of the series formula underscores the complex nature of the problem. The presence of nested trigonometric functions suggests that KAN has made an effort to approximate the behavior of the analytical solution, akin to that of a Fourier series. However, the R² value (0.20) suggests that KAN's predictions may not be capturing the overall trend of the data effectively.

Figure \ref{fig:Re_Ca}, Case 3 shows the challenges faced by KAN in this scenario. While the predicted temperature distribution (blue solid line) captures the general trend of the actual distribution (orange dashed line), there are significant deviations, particularly at the beginning and the inability to capture the high-frequency changes. 

We hypothesize that the observed discrepancies can be attributed to two main factors. Firstly, the complexity of the problem (its spatial and temporal variations, periodic boundary conditions, and interactions between multiple frequencies) highlights the difficulty for users to integrate domain knowledge into the KAN modeling process. Capturing such intricate behaviors requires selecting appropriate mathematical operators and functions that embody the relevant physical principles. This task can be challenging, and if the domain knowledge is not adequately incorporated, the model may struggle to accurately represent the system's true underlying behavior by the default algorithmic setting but require more fine adjustments from problem-specific insights. Secondly, KAN's symbolic regression tends to discover multiple sub-optimal solutions rather than an optimal formula that accurately reflects the system's true physical behavior. The algorithm struggles to revisit and thoroughly search the vast space of potential formula expressions once certain paths are taken. Instead of finding fundamentally better-fitting expressions, it often adds more terms to existing formulas to minimize prediction errors. This approach can result in overly complex formulas that overfit in some regions and underfit in others, compromising overall interpretability.

\subsubsection{Case Study 3*: Significant Improvement through Prior Knowledge Integration}

Through this transformation, we can shift the solving method from a series of $\sin(f(x))$ terms to a linear summation. This streamlines the training process, as a single layer of KAN network is needed for the solution. Furthermore, we leveraged prior knowledge to fix the boundary function of the KAN network as a linear function. This approach aligns the problem formulation more closely with KAN's underlying principles, particularly its ability to handle linear combinations effectively.

The modified approach has substantially enhanced the results. Prediction performance across the entire dataset has dramatically improved, with the \( R^2 \)  value increasing to 0.99.   An illustrative presentation of the predictive result is shown in Figure \ref{fig:Re_Ca}, case 3*. We also discovered a formula (for the complete expression, see\ref{Case Study Results}) that closely resembles those derived from physical equations:

\begin{equation}
HF(\tau) = \sum_{j=0}^{\infty} a_j\cdot t_a(\tau-j) +B,
\label{eq:HF_formula2}
\end{equation}



This improvement demonstrates the importance of aligning the problem formulation with the strengths of the machine learning model. By transforming the problem into a linear combination, we leveraged KAN's ability to handle such structures effectively. It also showed the value of incorporating domain knowledge into the model design and the ability of KAN to discover physically meaningful relationships when guided appropriately. 

While the initial results of Case 3 presented challenges, our subsequent refinement demonstrates KAN's potential to handle even highly complex, dynamic systems when appropriately guided by domain knowledge and mathematical insight.

\subsubsection{Case Study 4: Performance Comparison with MLPs}
\label{subsubsec:case4_analysis}

\begin{figure}[h]
    \centering
    \includegraphics[width=0.5\linewidth]{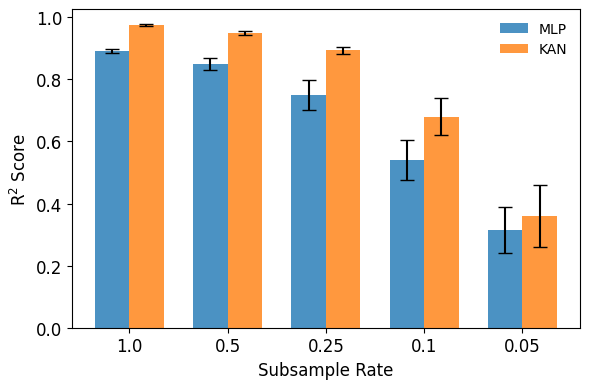}
    \caption{Robustness comparison against data sparseness between KAN and MLP based on the whole dataset with the availability of 100\%, 50\%, 25\%, 10\%, and 5\% respectively, averaged over 10 random runs. KAN shows better accuracy and robustness against data sparseness.}
    \label{fig:Re_Sp}
\end{figure}

Table \ref{tab:combined_results} summarizes the R² scores for both MLP and KAN across all tasks and subsample rates at each evaluation point for verifying their continuous learning ability. The complete dataset is also subsampled in different ratios to evaluate the robustness of MLP and KAN against the data sparseness, as the result presented in Figure \ref{fig:Re_Sp}. 

\begin{table}[h]
\centering
\caption{Continuous learning ability: R² Scores (mean ± standard deviation) for MLP and KAN averaged over 10 random runs. The data is split into three random subsets (tasks), and continuously train the model with a new task and validates its performance on all tasks.}
\label{tab:combined_results}
\resizebox{\textwidth}{!}{ 
\begin{tabular}{ccccccccccc}
\hline
\multirow{2}{*}{Model} & \multirow{2}{*}{\makecell{Subsample \\ Rate}} & \multicolumn{3}{c}{After Task 1} & \multicolumn{3}{c}{After Task 2} & \multicolumn{3}{c}{After Task 3} \\
\cmidrule(lr){3-5} \cmidrule(lr){6-8} \cmidrule(lr){9-11}
& & T1 & T2 & T3 & T1 & T2 & T3 & T1 & T2 & T3 \\
\hline
\multirow{4}{*}{MLP} & 100\% & 0.87 ± 0.02 & 0.82 ± 0.03 & \textbf{0.83 ± 0.02} & 0.87 ± 0.02 & 0.87 ± 0.02 & 0.85 ± 0.02 & 0.87 ± 0.01 & 0.85 ± 0.01 & 0.87 ± 0.01 \\
& 50\% & 0.75 ± 0.02 & 0.64 ± 0.03 & 0.63 ± 0.03 & 0.72 ± 0.03 & 0.80 ± 0.03 & 0.70 ± 0.04 & 0.76 ± 0.02 & 0.77 ± 0.02 & 0.82 ± 0.03 \\
& 25\% & 0.64 ± 0.09 & 0.46 ± 0.07 & 0.45 ± 0.07 & 0.60 ± 0.06 & 0.69 ± 0.05 & 0.53 ± 0.04 & 0.63 ± 0.03 & 0.62 ± 0.04 & 0.71 ± 0.03 \\
& 10\% & 0.52 ± 0.13 & \textbf{0.22 ± 0.06} & \textbf{0.21 ± 0.04} & \textbf{0.40 ± 0.09} & 0.55 ± 0.13 & \textbf{0.30 ± 0.07} & \textbf{0.40 ± 0.05} & \textbf{0.44 ± 0.10} & 0.57 ± 0.08 \\
\hline
\multirow{4}{*}{KAN} & 100\% & \textbf{0.96 ± 0.00} & \textbf{0.82 ± 0.02} & 0.82 ± 0.02 & \textbf{0.92 ± 0.00} & \textbf{0.94 ± 0.00} & \textbf{0.91 ± 0.00} & \textbf{0.93 ± 0.01} & \textbf{0.91 ± 0.01} & \textbf{0.94 ± 0.00} \\
& 50\% & \textbf{0.95 ± 0.01} & \textbf{0.89 ± 0.01} & \textbf{0.89 ± 0.01} & \textbf{0.91 ± 0.01} & \textbf{0.95 ± 0.01} & \textbf{0.91 ± 0.01} & \textbf{0.92 ± 0.00} & \textbf{0.92 ± 0.01} & \textbf{0.95 ± 0.01} \\
& 25\% & \textbf{0.92 ± 0.01} & \textbf{0.73 ± 0.04} & \textbf{0.75 ± 0.04} & \textbf{0.83 ± 0.02} & \textbf{0.94 ± 0.01} & \textbf{0.83 ± 0.01} & \textbf{0.86 ± 0.02} & \textbf{0.87 ± 0.02} & \textbf{0.94 ± 0.00} \\
& 10\% & \textbf{0.83 ± 0.03} & 0.14 ± 0.18 & 0.18 ± 0.15 & 0.31 ± 0.17 & \textbf{0.84 ± 0.05} & 0.20 ± 0.16 & 0.25 ± 0.17 & 0.36 ± 0.14 & \textbf{0.85 ± 0.04} \\
\hline
\end{tabular}
}
\vspace{10 pt} 
\raggedright 
\small{Note: Bold values indicate the highest mean R² score between MLP and KAN for each subsample rate and evaluation point.}
\end{table}

The comparative analysis of KAN and MLP across different subsample rates and learning tasks reveals that the KAN outperforms MLP in terms of R² scores across most subsample rates and tasks, demonstrating superior adaptability and performance. KAN shows promising stability in maintaining high performance across tasks. MLP, while generally showing lower R² scores compared to KAN, demonstrates a gradual improvement in performance across tasks, particularly evident in the 50\% and 25\% subsample rates. Interestingly, at the lowest subsample rate (10\%), both models struggle, but MLP shows slightly better consistency across tasks. KAN appears to be a superior choice when rapid adaptation and high performance are crucial, especially when working with moderate to abundant data. However, MLP might be preferable in scenarios with limited data or where gradual, cumulative learning is desired.

\begin{figure}[h]
    \centering
    \includegraphics[width=1\linewidth]{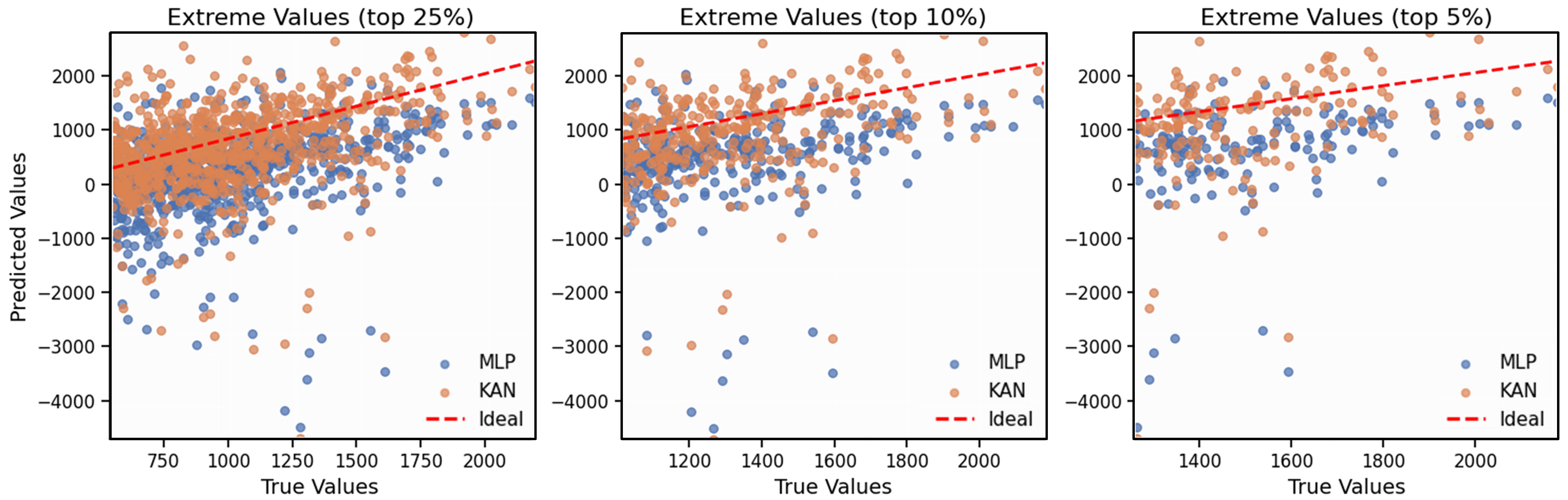}
    \caption{Performance comparison between KAN and MLP in capturing top 25\%, 10\%, and 5\% extreme values. KAN shows better accuracy with less underestimation of extreme values.}
    \label{fig:Re_Ex}
\end{figure}

For the robustness test, KAN shows superior adaptability than MLP in data-sparse environments, particularly at lower subsample rates. The reason could be due to its symbolic optimization nature. 
Furthermore, the analysis result of extreme value predictions, as shown in Figure \ref{fig:Re_Ex}, reveals KAN's superior performance in capturing extreme values, especially at higher percentiles.In this study, MLP has a tendency to underestimate extreme values, with this trend intensifying at higher extremities. This tendency can have significant implications in fields such as building physics, where underestimating peak energy demands or structural loads might lead to inadequate system design or safety margins. Therefore, KAN's enhanced performance in extreme value prediction may offer a more reliable tool in scenarios where accurate extremes are critical for system sizing. 

\section{Decision Tree: Model Selection}
\label{sec:Framework}

In this section, we summarize our previous results and key information on prior knowledge embedding, interpretability, continual learning, and robustness to data sparsity.

KANs can be used for predictions (for example, building energy load prediction \cite{zhang2021review}, PV systems generation forecasting \cite{kabilan2021short}, and indoor environment parameter prediction \cite{zhou2022realtime}), they have the potential for more interpretable models compared to MLPs, they also improve accuracy under limited data and enable better transfer learning. KANs can also be used for system diagnosis (for example, HVAC fault diagnosis \cite{taheri2021fault} and thermal bridge detection \cite{mayer2023deep}) because they give the benefit of interpretable methods that can capture temporal dependencies and dynamic behavior of faults. Moreover, KANs can be used to extract underlining knowledge, for example, areas like AI-assisted material design \cite{lu2024digital} and personalized thermal comfort modeling \cite{tartarini2022personal}.

\begin{figure}[htbp]
    \centering
    \includegraphics[width=1\linewidth]{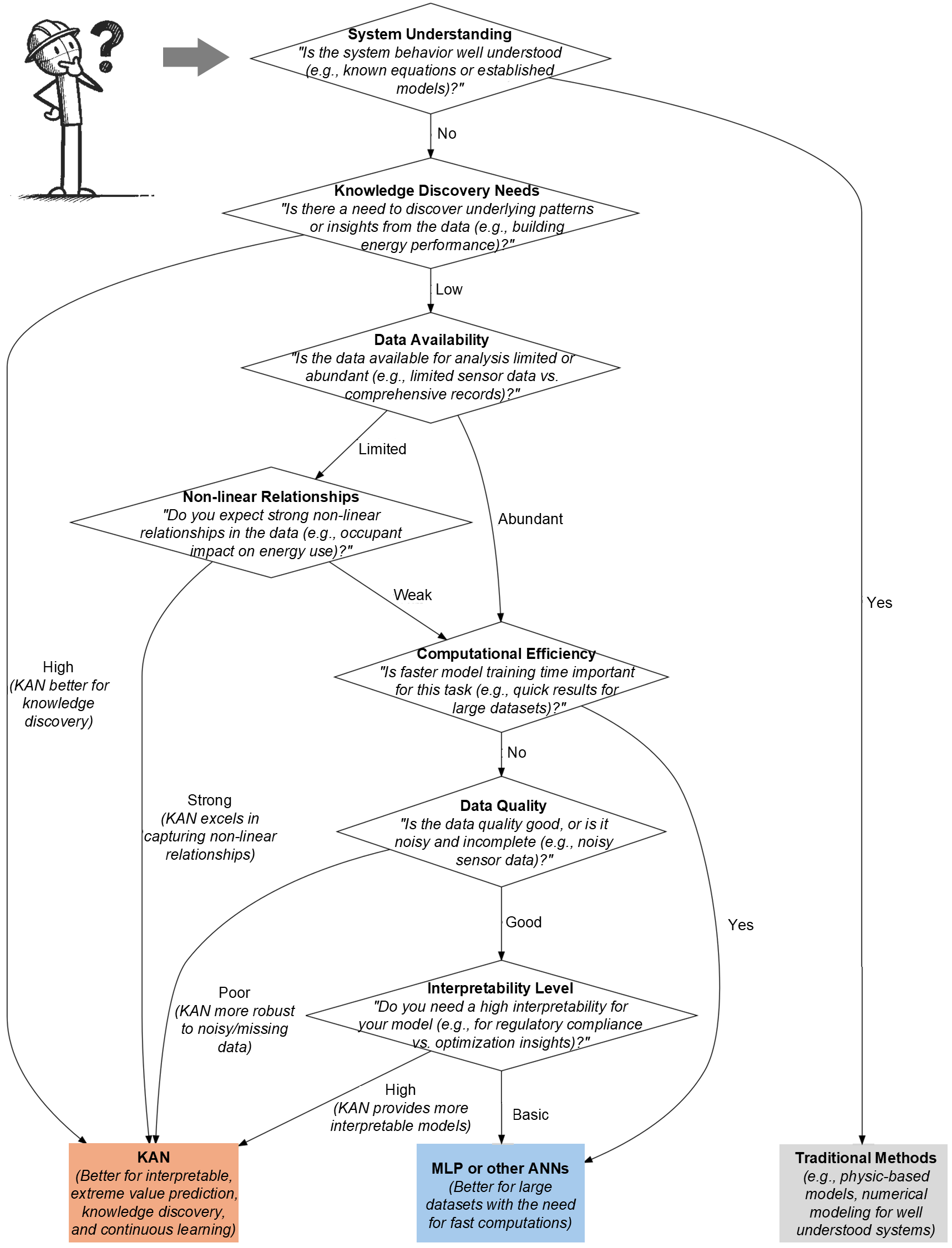}
    \caption{Decision tree for choosing between KAN and MLP in building physics applications.}
    \label{fig:Re_Tree}
\end{figure}

Based on these findings, we propose a decision tree to guide researchers in selecting the most appropriate ML tool for their specific challenges in building physics, as illustrated in Figure \ref{fig:Re_Tree}.

The key distinction we observe lies in the interaction patterns between users and these different ML approaches: Traditional MLPs and ML models primarily function as computational tools for prediction. The interaction framework with these models is typically characterized by a direct input-output mapping process. Users provide input data and receive predictive outputs, with minimal transparency regarding the internal mechanisms of the model. Although effective in various applications, this method may limit the depth of interpretability and understanding that users can derive from the model. In contrast, interpretable methods served as instruments for knowledge extraction to facilitate the identification of underlying relationships and patterns, which can be incorporated as constraints in further analytical processes. The interaction process with these methods is iterative, enabling continuous refinement and discovery of knowledge. This approach allows for cross-validation with existing domain knowledge and supports adaptive learning as new data becomes available, enhancing both the interpretability and applicability of the models in complex, evolving domains.

Underpinned by the decision-making process outlined in Figure \ref{fig:Re_Tree}, we hope that this study is an advancement in applying ML methods to complex physical systems in building science. By carefully formulating problems and incorporating prior knowledge, we can guide these methods towards solutions that are both accurate and physically interpretable, paving the way for more effective human-AI collaboration in building physics research and applications.

\section{Conclusion}
\label{sec:Conclusion}
This study demonstrates the potential of Kolmogorov–Arnold Networks for solving inverse problems in building physics, showcasing its capabilities in predictive modeling, knowledge discovery, and continuous learning. While KAN faces challenges in extrapolation and interpretability, its strengths in robustness against data sparseness, continuous learning, adaptability, and capturing extreme values make it a valuable tool for advancing building science. Lastly, to guide the appropriate use of machine learning tools in building physics, we present a decision tree to assist researchers in selecting the most suitable ML model for specific problems, considering factors such as data availability, problem complexity, and desired outcomes. 

\section{Appendix}
\label{sec:Appendix}
\appendix
\section{Implementation Details and Case Setup}
\label{app:implementation}

This appendix provides additional information on the implementation of the Kolmogorov–Arnold Networks (KAN) and the specific setup for each case study. The details presented here are crucial for increasing the reproducibility and for researchers who wish to build upon or validate our work.

\subsection{Case Study Setups}
\label{app:case_study_setups}

\subsubsection{Case Study 1: Steady-State Heat Conduction}
\label{app:case1_setup}

The temperature distribution in the wall under steady-state heat conduction can be determined by solving the heat equation as follows:

\begin{equation}
\label{steady conduction}
\frac{d}{dx} \left( k \frac{dT}{dx} \right) = 0
\end{equation}

When the temperatures on both sides of the wall are \(T_1\) and \(T_0\), respectively, and the wall thickness is \(L\), the temperature distribution is given by eq. \eqref{wall steady conduction}. We discretized the wall thickness into 100 equally spaced points, generating a total of 100 data points. The input feature was the normalized position (x/L) along the wall, while the output was the normalized temperature ((T-T\textsubscript{0})/(T\textsubscript{L}-T\textsubscript{0})). This normalization scheme was chosen to make the model more generalizable to walls of different thicknesses and temperature ranges.

The KAN for this case was configured with an input layer of 1 neuron, two hidden layers of 20 and 10 neurons respectively, and an output layer of 1 neuron. We used a diverse set of activation functions including 'x', 'sin', 'exp', 'log', and 'sqrt' to capture potential nonlinearities in the relationship. The data was split into 70\% training, 15\% validation, and 15\% test sets.

Hyperparameters were fine-tuned through a grid search process. The learning rate was set to 0.001 with an exponential decay (rate: 0.95, steps: 1000) to ensure stable convergence. We used a batch size of 32 and an L2 regularization coefficient of $1 \times 10^{-5}$ to prevent overfitting.

\subsubsection{Case Study 2: Transient Heat Conduction}
\label{app:case2_setup}

Soil can be considered as a semi-infinite solid. If a sudden change in conditions is applied at the surface, transient one-dimensional conduction will occur within the solid. It can be represented as follows:
\begin{equation}
\label{Transient Heat Conduction}
\frac{\partial t}{\partial \tau} = \alpha \frac{\partial^2 t}{\partial x^2}
\end{equation}
with the boundary conditions \(T(t) = T_0\) and \(T(x = 0, t \to \infty) = T_1\)

Introducing the composite variable of time and temperature, \(\eta = \frac{x}{(4 \alpha t)^{1/2}}\), into the eq. \eqref{Transient Heat Conduction}, we obtain the following eq. \eqref{Semi-Infinite Transient Heat Condcution}:

\begin{equation}
    \label{Semi-Infinite Transient Heat Condcution}
    \frac{T - T_0}{T_1 - T_0} = \left( \frac{2}{\pi^{1/2}} \right) \int_{0}^{\eta} \exp(-u^2) \, du = \operatorname{erf} \, \eta
\end{equation}

The transient heat conduction case required a more complex setup. We discretized the wall into 50 equally spaced points and considered 100 time steps over a 24-hour period, resulting in 5000 total data points. The input features were the normalized position (x/L) and normalized time (t/t\textsubscript{max}), with the output being the normalized temperature ((T-T\textsubscript{0})/(T\textsubscript{1}-T\textsubscript{0})).

To handle this increased complexity, we configured the KAN with an input layer of 2 neurons, three hidden layers of 30, 20, and 10 neurons respectively, and an output layer of 1 neuron. The activation functions were expanded to include 'erf' in addition to those used in Case 1, allowing the network to better approximate the analytical solution involving the complementary error function. We used a 60-20-20 split for training, validation, and test sets.

Hyperparameters were adjusted to account for the increased complexity. The learning rate was reduced to 0.0005 and employed a cosine decay schedule over 5000 steps. The batch size was increased to 64, and the L2 regularization coefficient was set to $5 \times 10^{-5}$.

\subsubsection{Case Study 3: Dynamic Heat Transfer with Periodic Boundary Conditions}
\label{app:case3_setup}

In the field of building physics, it is often necessary to solve dynamic heat transfer problems with periodic boundary conditions, especially calculating the heat flow transmitted to the interior through the building envelope in real time. This is crucial for the design and control of building air conditioning systems.

As for the envelope heat transfer, it can be simplified as a one-dimension dynamic heat transfer when the length was longer than width 8-10 times. The equation is as eq.\eqref{Transient Heat Conduction}. Apply the Laplace transform to the time-variable function \( t(x, \tau) \) in the Equation:
\begin{equation}
\label{L-transform}
L\{t(x, \tau)\}_s = \int_0^\infty t(x, \tau) e^{-s\tau} d\tau = T(x, s)
\end{equation}
By substituting eq. \eqref{L-transform} into Fourier's law, we can obtain eq. \eqref{Q_transform} :
\begin{equation}
\label{Q_transform}
Q(x, s) = -\lambda T'(x, s)
\end{equation}

A single-layer homogeneous plate with thickness \( l \), thermal conductivity \( \lambda \), and thermal diffusivity \( a \). The outdoor side temperature and heat flow are given by \( T(0, s) \) and \( Q(0, s) \). By substituting the derived expressions into the equations, we obtain the indoor side temperature and heat flow \( T(l, s) = Q(l, s) \):

\begin{equation}
\begin{bmatrix}
T(l, s) \\
Q(l, s)
\end{bmatrix}
=
\begin{bmatrix}
\cosh\left(\sqrt{\frac{s}{a}} \cdot l\right) & -\frac{\sinh\left(\sqrt{\frac{s}{a}} \cdot l\right)}{\lambda \sqrt{\frac{s}{a}}} \\
-\lambda \sqrt{\frac{s}{a}} \sinh\left(\sqrt{\frac{s}{a}} \cdot l\right) & \cosh\left(\sqrt{\frac{s}{a}} \cdot l\right)
\end{bmatrix}
\begin{bmatrix}
T(0, s) \\
Q(0, s)
\end{bmatrix}
\end{equation}

In a variable temperature environment, the temporal change can be represented using a Fourier series, an approximation for the temperature change over time \(\tau \) can be represented as follows, 
\begin{equation}
t_a(\tau) = t_0 + \sum_{n=1}^\infty A_n \sin(\omega_n \tau + \phi_n)
\end{equation}

where \( t_0 \) corresponds to the average temperatures. \( A_n \) is the amplitude of the \( n \)-th harmonic component of temperature in degrees Celsius. \( \phi_n \) is the phase shift of the \( n \)-th harmonic component in radians. \( \omega_n \) is the angular frequency of the \( n \)-th harmonic.

By applying the transfer function from eq.\eqref{L-transform}, we can compute the attenuation \( v_n \) of the wall to the outdoor air temperature, as well as the corresponding delay phase \( \psi_n \). Subsequently, we can determine the fluctuations in the indoor side wall temperature, which are influenced by various degrees of temperature fluctuations from the exterior. The formula for the indoor side temperature fluctuations is expressed as follows:

\begin{equation}
\Delta t(\tau) = \sum_{n=1}^N \frac{A_{n}}{v_n} \sin(\omega_n \tau + \phi_n - \psi_n)
\end{equation}

Then, using the average temperatures indoors \( t_{in} \)and outdoors \( t_o \), the wall's heat transfer coefficient \( K \), and the convection heat transfer coefficient of indoor side \( h_{\text{in}} \), the dynamic heat flow on the indoor side can be calculated.

\begin{equation}
HF(\tau) =  K\left[t_0 - t_{in} + \frac{h_{in}}{K} \ \sum_{n=1}^N \frac{A_{n}}{v_n} \sin(\omega_n \tau + \phi_n - \psi_n)\right]
\end{equation}

Outdoor temperature profiles can also be discretized for analysis. Then, by utilizing the residue theorem, we can solve for the response factors \( Y(j) \) to quantify the impact of each time node of outdoor temperature \( t_a(\tau-j) \) on heat flow, thereby determining the dynamic heat flow. The calculation can be represented by the following equation:

\begin{equation}
HG(\tau) = \sum_{j=0}^{\infty} Y(j) t_a(\tau-j) - K t_{in},
\end{equation}

The Response Factor Method involves applying the Laplace transform to the heat transfer differential equations, then solving for the impact of thermal perturbations at discrete times (i.e., response factors), and finally calculating the final results through superposition. This method is used to compute the heat flow from the exterior of a wall into the interior under the conditions of varying external temperatures, with the interior side temperature fixed at $26^\circ\mathrm{C}$. The expression for this is as follows, which consists entirely of linear additions and subtractions, similar to the expression form of KAN. This implies that shallow KAN networks might suffice to capture dynamic heat transfer characteristics during subsequent training processes.

\begin{equation}
HG(n) = \sum_{j=0}^{\infty} Y(j) t_z (n - j) - \sum_{j=0}^{\infty} Z(j) t_r (n - j)
\end{equation}

where, $Y(j)$ and $Z(j)$ are response factors, $t_z(n-j)$ is the external composite temperature with a delay of $j$ time units, and $t_r$ is the interior side temperature.

\subsubsection{Case Study 4: Performance Comparison with MLPs}
\label{app:case_study_4}

To further evaluate the performance of KAN in building physics applications, we conducted case study 4 to compare KAN with traditional MLPs. We implemented a feedforward neural network with an input layer of 7 neurons (corresponding to the input features), two hidden layers of 64 and 32 neurons respectively, and an output layer of 1 neuron. ReLU activation was used for hidden layers, and a linear activation for the output layer. The KAN architecture consisted of an input layer of 7 neurons, one hidden layer of 20 neurons, and an output layer of 1 neuron, with a grid size of 10 and k value of 10.

Both models were trained using the Adam optimizer for the MLP and L-BFGS for KAN, with MSE as the loss function. We implemented early stopping to prevent overfitting, with a patience of 10 epochs for MLP and 3 steps for KAN. The maximum number of epochs was set to 200 for MLP and 100 steps for KAN.

To evaluate model performance under varying data availability, we trained both models on progressively smaller subsets of the data, using subsample rates of 100\%, 50\%, 25\%, 10\%, and 5\%. For each subsample rate, we randomly subsampled the dataset, trained both models, and evaluated their performance on a separate test set. We recorded R² scores to assess prediction accuracy.

We designed a sequential learning scenario with three tasks to simulate a continuous learning environment. The data for all three tasks was normalized using only the statistics from Task 1, simulating a real-world scenario where future data distributions may be unknown. For each subsample rate (100\%, 50\%, 25\%, and 10\%), we trained the models sequentially on the three tasks. After training on each task, we evaluated the models' performance on all tasks completed up to that point. This approach allowed us to assess how well the models adapted to new information and maintained performance on earlier tasks under varying data availability conditions. The entire process was repeated 10 times with different random seeds to ensure the statistical reliability of our results. For each subsample rate, the experiment proceeded as follows: First, we normalized the data using only Task 1's statistics, applying this normalization to all three tasks. This approach more closely mimics real-world scenarios where future data distributions may be unknown. We then continuously trained the models on each task in sequence and evaluated their performance on all three tasks after each training phase. This process was repeated 10 times with different random seeds to ensure statistical robustness. This experimental design allows us to observe how each model adapts to new tasks and retains knowledge of previously learned tasks under different data availability conditions, while also accounting for potential distributional shifts between tasks.

To assess the models' ability to predict rare or extreme events, we focused on their performance in capturing the top 25\%, 10\%, and 5\% of the heat flow values in the dataset. For each threshold, we identified data points above it and evaluated model predictions for these points.

In addition to the metrics mentioned in Section \ref{app:evaluation_metrics}, we included Mean Absolute Percentage Error (MAPE), Symmetric Mean Absolute Percentage Error (SMAPE), and the Pearson correlation coefficient for a more comprehensive evaluation. We visualized results using bar plots for R² score comparisons across different subsample rates, scatter plots for actual vs. predicted values in extreme value analysis, and line plots to show continuous learning performance across tasks.

\subsection{Case Study Results}
\label{Case Study Results}
The specific expressions of the formula\eqref{eq:HF_formula1}  and\eqref{eq:HF_formula2} in the main text are presented below with their respective coefficients: 

\begin{align}
HF &= 3.11\sin(0.98x_1 - 4.21) + 0.57\sin(0.94x_{10} + 1.56) + 6.03\sin(0.48x_{11} - 1.24) \nonumber \\
&\quad  - 0.94\sin(0.64x_{12} + 8.04) - 0.57\sin(0.69x_{13} + 8.00) + 3.30\sin(0.55x_{14} - 1.30) \nonumber \\
&\quad - 0.47\sin(0.89x_{15} - 7.82) - 0.38\sin(0.84x_{16} + 4.77) + 3.39\sin(0.56x_{17} - 1.30)  \nonumber \\
&\quad + 5.38\sin(0.48x_{18} - 1.25) - 8.68\sin(0.44x_{19} + 1.93) - 2.92\sin(1.14x_2 + 4.91) \nonumber \\
&\quad  - 0.19\sin(0.73x_{20} - 4.62) + 2.64\sin(0.54x_{21} - 1.29) - 2.07\sin(1.96x_{22} - 8.60)  \nonumber \\
&\quad   - 2.07\sin(1.41x_{23} + 4.73) - 6.79\sin(0.81x_{24} + 5.18) - 0.85\sin(1.17x_3 - 1.68)  \nonumber \\
&\quad   + 1.32\sin(1.88x_4 + 9.15) + 3.11\sin(0.56x_5 - 1.30) + 2.55\sin(0.57x_6 - 1.32) \nonumber \\
&\quad   - 1.13\sin(1.39x_7 + 4.43) - 5.94\sin(0.87x_8 + 5.13) - 1.70\sin(0.60x_9 + 8.07) + 28.72
\end{align}

\begin{align}
HF &= -0.19x_1 + 1.04x_{10} + 0.94x_{11} + 0.75x_{12} + 0.66x_{13} + 0.57x_{14} + 0.47x_{15} + 0.47x_{16}  \nonumber \\  
&\quad  + 0.38x_{17} + 0.28x_{18} + 0.28x_{19} + 0.19x_{20} + 0.19x_{21} + 0.19x_{22} + 0.19x_{23} + 0.19x_{24} \nonumber \\ 
&\quad + 0.47x_{25} + 0.09x_3  + 0.47x_4 + 1.04x_5 + 1.32x_6 + 1.32x_7 + 1.22x_8 + 1.13x_9 + 9.74
\end{align}

\subsection{KAN Implementation}
\label{app:kan_implementation}

The KAN architecture was designed to be flexible, adapting to the complexity of each case study. The input layer varied from 1 to 2 neurons, corresponding to the spatial and temporal dimensions of our problems. The hidden layers were structured with variable width and depth, determined through extensive hyperparameter tuning to optimize performance for each specific case. This adaptive approach allowed us to balance between model expressiveness and computational efficiency. The output layer consistently comprised a single neuron, representing the temperature prediction. A key feature of our KAN implementation was the use of a combination of standard activation functions (such as ReLU and tanh) and symbolic functions, enabling the network to capture both simple and complex relationships in the data.

\subsubsection{Training Process}
\label{app:training_process}

The training process was carefully designed to ensure optimal performance and convergence. We employed the Adam optimizer, known for its efficiency in training deep neural networks, coupled with a learning rate scheduling mechanism to fine-tune the optimization process as training progressed. The loss function was defined as the Mean Squared Error (MSE), providing a clear measure of prediction accuracy. To prevent overfitting, we implemented L2 regularization and early stopping. The training duration was variable, with an early stopping patience of 100 epochs, allowing the model sufficient time to converge while preventing unnecessary computational overhead. 

\subsubsection{Symbolic Optimization}
\label{app:symbolic_optimization}

The symbolic optimization component of KAN, crucial for discovering interpretable equations, was implemented using a Genetic Programming approach. This algorithm evolves a population of potential solutions over multiple generations, progressively improving their fitness. We defined the fitness function as a combination of prediction accuracy and formula simplicity, striking a balance between performance and interpretability. The population size was set to 1000, evolving over 100 generations. Mutation and crossover rates were set to 0.1 and 0.7 respectively, values determined through preliminary experiments to provide a good balance between exploration and exploitation in the search space.

\subsection{Data Preprocessing and Normalization}
\label{app:data_preprocessing}

Consistent data preprocessing and normalization techniques were applied across all case studies to ensure comparability of results. Input features underwent min-max normalization to the range [0, 1], preserving the relative relationships while bringing all inputs to a common scale. Output temperatures were subjected to z-score normalization, centering the data around zero with unit variance. This approach helps in stabilizing the training process and often leads to faster convergence. Prior to splitting the datasets into training, validation, and test sets, we applied random shuffling to ensure that each set was representative of the entire data distribution, mitigating potential biases from the data generation process.

\subsection{Evaluation Metrics}
\label{app:evaluation_metrics}

To assess the performance of KAN across different scenarios, we employed the R-squared (R²) metric to quantify the proportion of variance in the dependent variable predictable from the independent variables. Additionally, we considered the formula complexity, measured by the number of terms in the discovered equation, as a proxy for the interpretability of the KAN's output. This multi-faceted evaluation approach enabled us to balance accuracy, generalizability, and interpretability in our assessment.

\subsection{Reproducibility}
\label{app:reproducibility}

Recognizing the importance of reproducibility in scientific research, we took several measures to ensure our results can be replicated and built upon. We have made our complete codebase, including data generation scripts, KAN implementation, and analysis notebooks, available in a public GitHub repository (\url{https://github.com/chenxiachan/KAN_inversed_problem}). This repository includes detailed documentation on the setup process, dependencies, and step-by-step instructions for running the experiments. We encourage researchers to utilize these resources to validate our findings and extend this work to new domains within building physics and beyond.


\section*{Acknowledgments}
This study was supported by the China Scholarship Council (NO. 
202306320344), and the Center for the Built Environment at the University of California, Berkeley. We acknowledge the use of <ChatGPT-4 (Open AI, \url{https://chat.openai.com/})> to identify improvements in the writing style, and create the cartoon image (DALL·E 3, Open AI) included in Figure \ref{fig:Re_Tree} (top-left) for visual illustration purposes. The underlying ideas, methodologies, data analysis, and interpretation of results were developed independently by the authors. All authors guarantee the originality of the content and are responsible for the analysis results presented in this paper.

\section*{Author contributions}
\textbf{Xia Chen}: Conceptualization, Methodology, Software, Validation, Investigation, Resources, Writing - Original Draft, Review, Visualization, Project administration; 
\textbf{Guoquan Lv}: Conceptualization, Methodology, Software, Validation, Investigation, Resources, Writing - Original Draft, Review, Visualization, Project administration; 
\textbf{Xinwei Zhuang}: Investigation, Review, Visualization; 
\textbf{Carlos Duarte}: Supervision, Validation, Review; 
\textbf{Stefano Schiavon}: Supervision, Validation, Review, Resources; 
\textbf{Philipp Geyer}: Supervision, Validation, Review, Resources;

\bibliographystyle{unsrt}  
\bibliography{references}

\end{document}